\documentclass[]{paxini}

\usepackage{wrapfig}
\usepackage{tabularx}
\usepackage{textcomp}
\usepackage{stfloats}
\usepackage{url}
\usepackage{verbatim}
\usepackage{graphicx}
\usepackage{titlesec}
\usepackage{tocloft}
\usepackage{adjustbox}
\usepackage{multirow}
\usepackage{pifont}
\usepackage{tikz}
\usepackage{comment}
\usepackage{amsmath,amssymb} 
\usepackage{colortbl}  
\usepackage{color}
\usepackage{booktabs} 
\usepackage{hyperref}
\usepackage{graphicx}    
\usepackage{subcaption} 
\usepackage{booktabs}
\usepackage{amsmath} 
\usepackage{multirow} 
\usepackage{booktabs} 
\usepackage{subcaption} 
\RequirePackage{xspace}
\makeatletter
\DeclareRobustCommand\onedot{\futurelet\@let@token\@onedot}
\def\@onedot{\ifx\@let@token.\else.\null\fi\xspace}

\makeatother

\usepackage{makecell}

\definecolor{adptorange}{RGB}{248, 205, 172}
\definecolor{cmpblue}{RGB}{189, 215, 238}
\definecolor{cmpblue}{RGB}{189, 215, 238}

\definecolor{our_red}{RGB}{232,157,160}
\definecolor{our_blue}{RGB}{136,206,230}
\definecolor{our_orange}{RGB}{246,200,168}
\definecolor{our_green}{RGB}{178,211,164}

\definecolor{attn_code0}{RGB}{247,215,200}
\definecolor{attn_code1}{RGB}{238,169,139}
\definecolor{mlp_code0}{RGB}{204,201,221}
\definecolor{mlp_code1}{RGB}{102,95,153}

\definecolor{token_blue}{RGB}{84, 120, 140}

\definecolor{myMagenta}{rgb}{0.9,0,0.4}

\usepackage{pifont}       
\usepackage{bbding}       
\usepackage{fontawesome}
\usepackage{xspace}

\usepackage{float}

\newlength\savewidth

\newcolumntype{x}[1]{>{\centering\arraybackslash}p{#1pt}}
\newcolumntype{y}[1]{>{\raggedright\arraybackslash}p{#1pt}}
\newcolumntype{z}[1]{>{\raggedleft\arraybackslash}p{#1pt}}

\renewcommand{\paragraph}[1]{\vspace{1mm}\noindent\textbf{#1}}
\usepackage{colortbl}
\usepackage{xcolor}
\usepackage{wrapfig}

\renewcommand{\paragraph}[1]{\vspace{1.25mm}\noindent\textbf{#1}}

\usepackage{algorithm}
\usepackage{listings}

\definecolor{codeblue}{rgb}{0.25, 0.5, 0.5}
\definecolor{codekw}{rgb}{0.35, 0.35, 0.75}
\lstdefinestyle{Pytorch}{
    language = Python,
    backgroundcolor = \color{white},
    basicstyle = \fontsize{9pt}{8pt}\selectfont\ttfamily\bfseries,
    columns = fullflexible,
    aboveskip=1pt,
    belowskip=1pt,
    breaklines = true,
    captionpos = b,
    commentstyle = \color{codeblue},
    keywordstyle = \color{codekw},
}

\definecolor{green}{HTML}{009000}
\definecolor{red}{HTML}{ea4335}

\title{RoboPaint: From Human Demonstration to Any Robot and Any View}

\author[1]{Jiacheng Fan}
\author[3, \ddagger]{Zhiyue Zhao}
\author[1]{Yiqian Zhang}
\author[1]{Chao Chen}
\author[1]{Peide Wang}
\author[1]{Hengdi Zhang}
\author[1, 2, \dagger]{Zhengxue Cheng}

\affiliation[1]{Paxini Tech.}
\affiliation[2]{Shanghai Jiao Tong University}
\affiliation[3]{Zhejiang University}

\contribution[\ddagger]{Work done during an internship at Paxini Tech.}
\contribution[\dagger]{Corresponding author}

\abstract{

Acquiring large-scale, high-fidelity robot demonstration data remains a critical bottleneck for scaling Vision-Language-Action (VLA) models in dexterous manipulation. We propose a Real-Sim-Real data collection and data editing pipeline that transforms human demonstrations into robot-executable, environment-specific training data without direct robot teleoperation. Standardized data collection rooms are built to capture multimodal human demonstrations (synchronized 3 RGB-D videos, 11 RGB videos, 29-DoF glove joint angles, and 14-channel tactile signals). Based on these human demonstrations, we introduce a tactile-aware retargeting method that maps human hand states to robot dex-hand states via geometry and force-guided optimization. Then the retargeted robot trajectories are rendered in a photorealistic Isaac Sim environment to build robot training data. Real world experiments have demonstrated: (1) The retargeted dex-hand trajectories achieve an 84\% success rate across 10 diverse object manipulation tasks. (2) VLA policies (Pi0.5) trained exclusively on our generated data achieve 80\% average success rate on three representative tasks, i.e., pick-and-place, pushing and pouring. To conclude, robot training data can be efficiently "painted" from human demonstrations using our real-sim-real data pipeline. We offer a scalable, cost-effective alternative to teleoperation with minimal performance loss for complex dexterous manipulation.
}

\date{\today}

\begin{document}
\thispagestyle{firstheader}
\maketitle
\pagestyle{empty}

\section{Introduction} \label{sec:introduction}
Recently, researchers have been working toward building general-purpose robotic agents using large Vision-Language-Action (VLA) models. Much like the evolution of Large Language Model or Vision Model, the performance of robotic systems is fundamentally tied to the scale, diversity, and quality of the underlying training data. High-fidelity demonstrations provide the essential mapping between multi-modal perceptions and precise motor commands, helping robots to learn complex manipulation skills. As robotics moves from simple pick-and-place tasks to more delicate, contact-heavy tasks, the demand for high-quality data that captures both precise motion and physical interaction has become increasingly critical.

Despite its importance, collecting high-quality robot data remains a significant bottleneck. Traditional teleoperation systems (\cite{1embodimentcollaboration2025openxembodimentroboticlearning} \cite{2gao2024efficientdatacollectionrobotic} \cite{3black2024pi0visionlanguageactionflowmodel}) offer high-fidelity control but suffer from limited scalability and high hardware costs, often requiring expert operators and physical robot platforms for every minute of collected data. To reduce the cost, handheld interfaces such as UMI (\cite{14chi2024universalmanipulationinterfaceinthewild}) have been introduced. But, this kind of device sacrifices the dexterity required for complex multi-fingered manipulation, which limits the data to only train simple gripper-style end-effectors. Meanwhile, many researchers use passive human videos from the web (\cite{mandlekar2023mimicgendatagenerationscalable}) to generate robot demonstration.  However, this approach faces a major challenge known as the "embodiment gap." This gap arises from structural and kinematic differences between human and robot hands, and the lack of tactile data in videos, which together hinder precise skill transfer.


To bridge these gaps, we propose \emph{\textbf{RoboPaint}}, a novel Real-Sim-Real data pipeline designed to scale high-fidelity dexterous manipulation datasets. Our approach begins with a high-precision Data-Acquisition Room where human operators, equipped with instrumented gloves, perform complex tasks. By capturing synchronized multi-view RGB-D video, joint kinematics, and tactile pressure maps, we record a complete "physical footprint" of human intent. To overcome the embodiment gap, we introduce a Dex-Tactile joint retargeting method that optimizes for both kinematic alignment and contact consistency in 3D space. Furthermore, we leverage 3D Gaussian Splatting (3DGS) to reconstruct photorealistic digital twins of the deployment environment. This allows us to "paint" human demonstrations onto any target robot embodiment within a high-fidelity simulation, generating massive-scale, multi-modal datasets that are ready for real-world VLA training.


\begin{figure*}[t!] 
    \centering
    \includegraphics[width=1\linewidth]{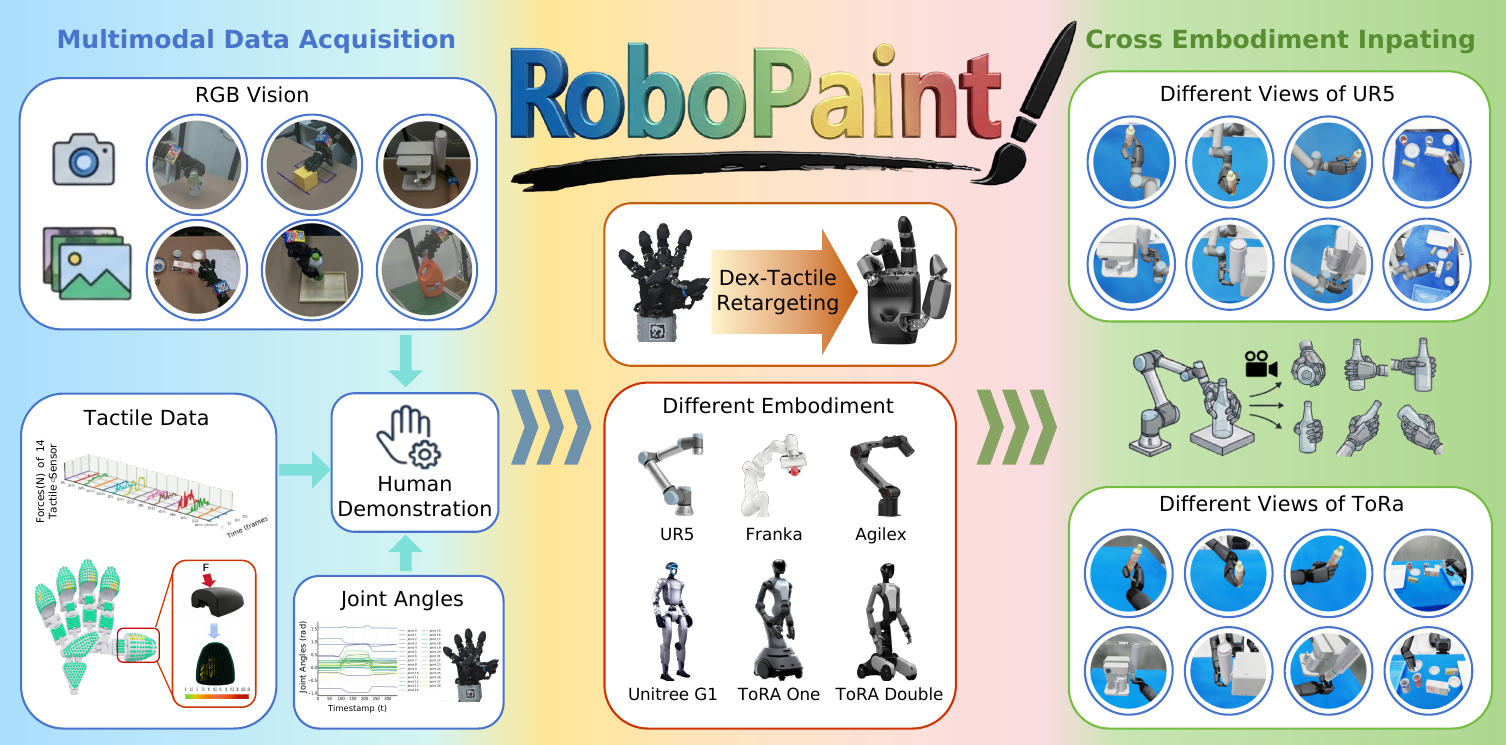}
    \caption{RoboPaint data pipeline. Our data pipeline can paint robot demonstration from multimodal data collected from human demonstration. The cross-embodiment problem between human and robot is resolved by our Dex-Tactile retargeting method.}
    \label{fig:teaser}
\end{figure*}



Our main contributions are summarized as follows:
\begin{itemize}
    \item We developed a high-fidelity multimodal data acquisition system capable of synchronously recording 11 channels of RGB visual signals (each has the resolution of $1200\times1920$), 3 channels of RGB-D multi-modal signals (each has the resolution of $720\times1280$), 15 channels of tactile signals (in total resolution of $3465\times3$), and 29 channels of proprioceptive joint signals (in total of $29\times1$).
    \item We propose \emph{\textbf{RoboPaint}}, a comprehensive data processing pipeline that integrates pose estimation, dex-tactile retargeting, and static scene construction to transform human demonstrations into robot-executable data. Our data processing toolkit is released: \url{https://github.com/px-DataCollection/px_omnisharing_dataprocess_kit}.
    \item Extensive result demonstrate that \emph{\textbf{RoboPaint}} achieves accurate modeling with an average contact error of only 3.86 mm and enables real-world robot motion replay with a success rate of 84\%. Furthermore, VLA models trained on our "painted" data achieved an overall manipulation success rate of 80\%. 
\end{itemize}

\section{Related Works}
\paragraph{Robot Data Collection.}
A variety of approaches have been developed for robot data collection, including leader-follower teleoperation systems such as ALOHA (\cite{5zhao2023learningfinegrainedbimanualmanipulation} \cite{6aloha2team2024aloha2enhancedlowcost}) and GELLO (\cite{7wu2024gello}), as well as interfaces based on VR/AR controllers (\cite{8jang2022bc} \cite{9khazatsky2024droid}), motion-capture systems (\cite{10bu2025agibot}), Project Aria (\cite{11kareer2025egomimic}), and smartphones (\cite{12wu2024tidybot++} \cite{13mandlekar2021matterslearningofflinehuman}). These methods all rely on real robot tele-operation, which, as mentioned before, imposes limitations in terms of scalability, cost, and data diversity. Alternative approaches, such as the UMI gripper (\cite{14chi2024universalmanipulationinterfaceinthewild}) and DOBB-E (\cite{15shafiullah2023bringingrobotshome}), enable data collection using handheld devices without full robotic platforms. However, they lack the dexterity required for fine-grained manipulation tasks, limiting their applicability to more complex scenarios. In contrast, our data collection setup preserves the high dexterity of human manipulation while simultaneously capturing high-quality visual and tactile signals, providing a more reliable avenue for constructing high-quality embodied datasets.

\paragraph{Human-Robot Visual Transfer.}
A key challenge in leveraging human demonstrations for robot learning lies in the embodiment gap: human and robot morphologies differ significantly in appearance, making raw human data unsuitable for direct use in training embodied policies. To mitigate this, prior work has explored visual-domain adaptation through image editing or visual domain normalization. For instance, RoviAug (\cite{16chen2024roviaugrobotviewpointaugmentation}) removes the original robot from each image and re-renders a target robot in the same pose before compositing it into the scene. Shadow (\cite{17lepert2025shadowleveragingsegmentationmasks}) replaces all robot pixels with segmentation masks during both training and inference to enforce a consistent input distribution. Similar strategies have been applied to human-to-robot transfer: EgoMimic (\cite{11kareer2025egomimic}) replaces human arms with black mask and overlays red line segments to mimic robotic arm trajectories. AR2-D2 (\cite{18peterson2010r2}) combines image inpainting with AR(Augmented Reality) rendering to structurally transform human demonstrations. Although these methods improve compatibility, the mismatch between human and robot remains. We align human motions with target robot embodiments directly in 3D space and render into 2D images, which helps producing more physically plausible and visually consistent data.

\paragraph{Learning on Human Data.}
Many studies have attempted to use human video data (\cite{19bahl2022humantorobotimitationwild} \cite{20bharadhwaj2024towards} \cite{21wang2023mimicplaylonghorizonimitationlearning} \cite{22xu2023xskill}), such as YouTube videos, to improve the generalization capability of robot policies. Some methods (\cite{8jang2022bc} \cite{23jain2024vid2robotendtoendvideoconditionedpolicy}) rely on paired human-robot data to bridge this gap. Others (\cite{24zhu2025visionbasedmanipulationsinglehuman}), estimate object poses from video demonstrations and learn robot policies based on object trajectories, while methods (\cite{25heppert2024ditto} \cite{26hsu2025spot}) track human body points to learn robot motion strategies. However, the structure of human hand and robot dex-hand are different, which hinders direct robot policy training. To address this challenge, we introduce a Dex-Tactile joint retargeting method that jointly optimizes kinematic alignment and tactile contact consistency, enabling precise, physics-aware transfer of human manipulation skills to dexterous robot embodiments.

\section{Method}

\begin{figure*}[t]
    \centering
    \includegraphics[width=1.0\textwidth]{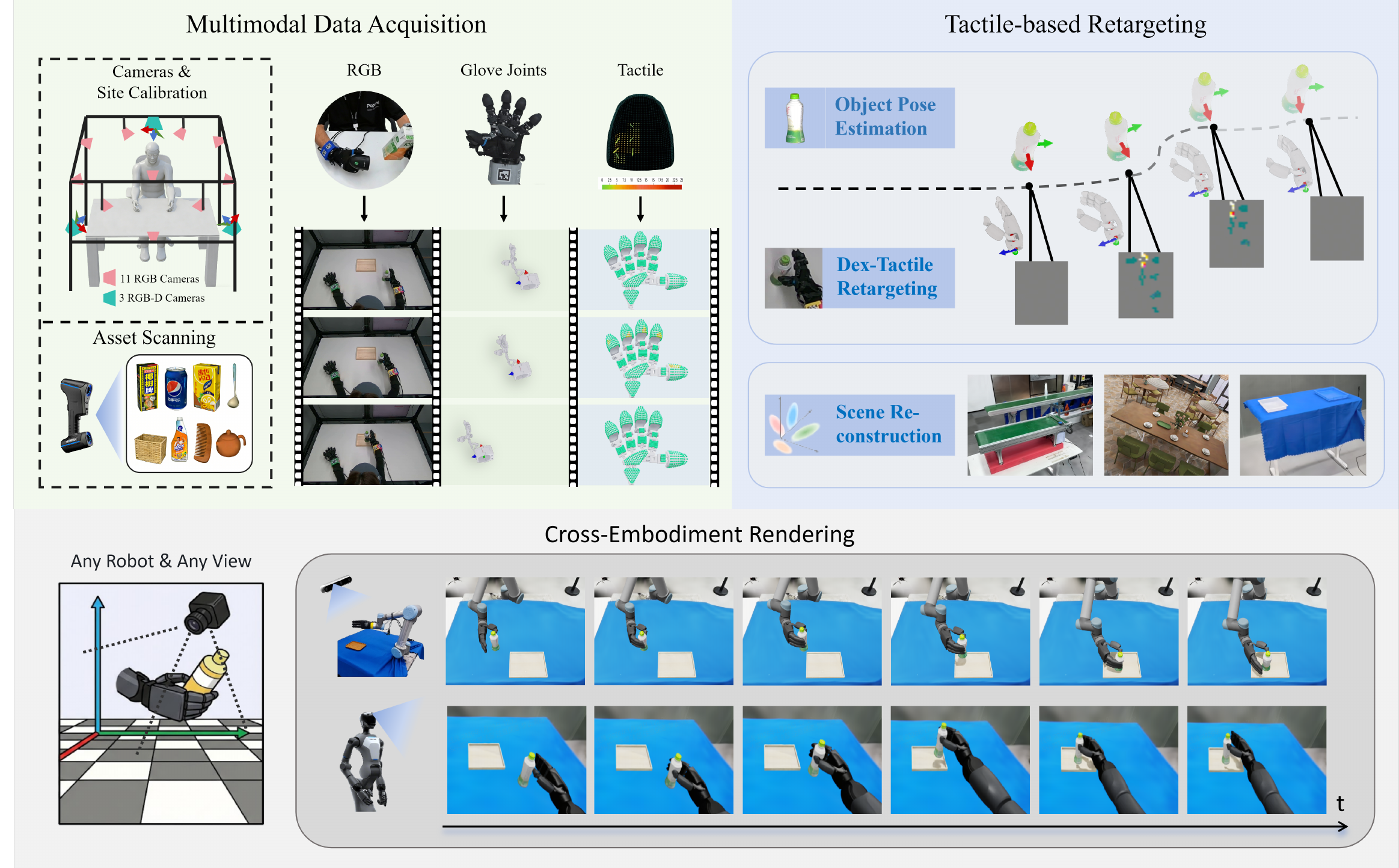}
    \caption{System Overview of the Real-Sim-Real Pipeline. Our framework begins with high-precision human data collection using instrumented gloves under our multiview Data-Acquisition-Room. The raw data is then processed through object pose estimation and Dex-Tactile joint retargeting to obtain the movement of target robot embodiment and object. The deployment scene is then reconstructed via 3D Gaussian Splatting and imported into simulation environment. Finally, we drive robot and objects accordingly and record robot demonstration from arbitrary view.}
    \label{fig:overview}
\end{figure*}

As illustrated in Fig.\ref{fig:overview}, our Real-Sim-Real data pipeline consists of three key stages. First, high-fidelity human demonstrations are captured in the Data-Acquisition-Rooms using 11 synchronized of RGB cameras and 3 synchronized RGB-D cameras. Second, 6D poses of the human wrists and manipulated objects are estimated. Based on glove joint angles, we develop a Dex-Tactile retargeting algorithm to map the human hand states to feasible target robot dex-hand states. Concurrently, the deployment scene is reconstructed using metrically aligned 3D Gaussian Splatting to enable photorealistic simulation. Third, the retargeted trajectories are transformed into the robot base coordinate frame and replayed in Isaac Sim, where they are rendered against the reconstructed 3D Gaussian Splatting background. Finally, the rendered visual observations, retargeted tactile signals, and robot action-state sequences are temporally aligned and packaged into multimodal datasets for training Vision-Language-Action (VLA) models.


\subsection{Human Data Acquisition}

To support large-scale, high-fidelity robot learning from human demonstrations, we have constructed many dedicated Data-Acquisition-Rooms, each equipped with a multi-camera array comprising both high-resolution RGB cameras and calibrated RGB-D sensors (Intel RealSense D455). All cameras are rigidly mounted and pre-calibrated with respect to the table world coordinate, enabling accurate 3D reconstruction and cross-view alignment of manipulation trajectories. In addition to constructing dedicated Data-Acquisition-Rooms, we have also built a structured object asset library to streamline asset management and reuse. Prior to data collection, all manipulated objects are scanned using a structured-light 3D scanner to generate high-quality digital assets, which are subsequently registered into the scene and used for downstream pose estimation and simulation.

During data collection, human operators perform a diverse repertoire of dexterous manipulation tasks such as grasping, lifting, pouring, insertion, and tool use while wearing custom-built instrumented gloves. These gloves integrate two complementary sensing modalities. High-precision magnetic rotary encoders are embedded at each finger joint to capture articulation angles with sub-degree resolution, and an array of Hall-effect-based tactile sensors is distributed across the palmar and fingertip regions to measure normal contact forces. Both sensor streams are sampled at high frequency and are hardware-synchronized with the visual system.

Concurrently, the multi-camera system records synchronized RGB and depth video streams at 30,Hz. A centralized time server ensures microsecond-level alignment across all modalities via hardware triggers and software timestamp interpolation. The resulting multimodal data, comprising RGB/RGB-D video, glove joint kinematics, and spatiotemporal tactile maps, are packaged into a unified, hierarchical HDF5 dataset with the following structure per demonstration sequence of duration $T$:
\begin{align}
    I_{i} &= \{\mathrm{img}_{\text{human}}^{t}\}_{t=1}^T \\
    J_{\text{Glove}} &= \{j_{\text{Glove}}^{t}\}_{t=1}^T \\
    \Gamma_{\text{Glove}} &= \{\gamma_{\text{Glove}}^{t}\}_{t=1}^T, \quad \gamma_{\text{Glove}}^{t} = \{f_i\}_{i=1}^{M} 
\end{align}
where, $I_i$ denotes the image stream from the $i$-th camera. $\mathrm{img}_{\text{human}}^{t} \in \mathbb{R}^{H \times W \times 3}$ is the RGB frame at timestamp $t$. $J_{\text{Glove}}$ captures hand kinematics via 20-dimensional joint angle vectors (5 joints across 4 fingers), while $\Gamma_{\text{Glove}}$ encodes the spatial pressure distribution. Specifically, $\gamma_{\text{Glove}}^{t} \in \mathbb{R}^{M}$ contains force readings from $M=32$ tactile sensors normalized to $[0,1]$. This dataset serves as the foundational input for our cross-embodiment retargeting pipeline, enabling faithful transfer of dexterous skills to robotic platforms.

This rich, temporally aligned, and geometrically grounded dataset provides a comprehensive record of human intent, hand morphology, and physical interaction. It serves as the foundational input for our cross-embodiment retargeting pipeline, enabling faithful transfer of dexterous skills to robotic platforms with dissimilar kinematic structures.

\subsection{Cross-Embodiment Modeling}

\subsubsection{Pose Estimation}
Based on the collected raw data, we estimate 6D poses of both human wrists and manipulated objects in the RGB-D camera coordinate. To track wrist motion robustly, we attach ArUco markers to custom wristbands worn by the operator. The pose of each marker is recovered using a sub-pixel accurate ArUco detector, yielding temporally smooth trajectories even under partial occlusion. For object pose estimation, we employ FoundationPose (\cite{wen2024foundationposeunified6dpose}) to achieve metrically accurate 6D object localization under varying lighting and clutter.

Let $P_{\text{Glove}} = \{p_{\text{Glove}}^{t}\}_{t=1}^T$ denote the estimated 4$\times$4 homogeneous pose matrices of the right and left wrists, respectively. Similarly, the object poses are denoted as $P_{\text{Object}} = \{p_{\text{Object}}^{t}\}_{t=1}^T$. All poses are expressed in the RGB-D camera coordinate and serve as inputs to the retargeting module.

\subsubsection{Dex-Tactile Retargeting}

Due to significant morphological differences between human hands and robotic dexterous hands (e.g., finger length, joint limits, actuation topology), raw glove data cannot be directly executed on the robot. Inspired by recent teleoperation frameworks (\cite{27qin2023anyteleop}), we design a physics-aware retargeting pipeline that maps human hand states to feasible robot dexterous hand states. In contrast to prior approaches that solely enforce geometric alignment between end-effectors, our method explicitly incorporates tactile feedback as an additional constraint during retargeting.

Specifically, given the glove inputs $(J_{\text{Glove}}, P_{\text{Glove}}, \Gamma_{\text{Glove}})$, our method optimizes for the target Paxini DexH13 states $(J_{\text{Dex}}, P_{\text{Dex}}, \Gamma_{\text{Dex}})$ by minimizing both kinematic and tactile constraints. Formally, at time stamp $t$, the retargeting minimizes the following energy function:
\begin{equation}
    \mathcal{L} = \mathcal{L}_{\text{kin}} + \mathcal{L}_{\text{tac}}
\end{equation}
The kinematic term $\mathcal{L}_{\text{kin}}$ is defined to align the spatial configuration:
\begin{equation}
    \mathcal{L}_{\text{kin}} = \frac{1}{N} \sum_{i=1}^{N} \left( \lambda_{\text{pos}} \| \mathbf{p}^{\text{Glove}}_i - \mathbf{p}^{\text{Dex}}_i \|_2 + \lambda_{\text{dir}} \| \mathbf{d}^{\text{Glove}}_i - \mathbf{d}^{\text{Dex}}_i \|_2 \right)
\end{equation}
Complementarily, the tactile term $\mathcal{L}_{\text{tac}}$ enforces contact consistency:
\begin{equation}
    \mathcal{L}_{\text{tac}} = \frac{1}{M} \sum_{j=1}^{M} w_{\mathbf{g}_j} \| \mathbf{g}_j - \mathrm{NN}_{\text{Dex}}(\mathbf{g}_j) \|_2
\end{equation}
where, $\mathbf{p}^{(\cdot)}_i$ and $\mathbf{d}^{(\cdot)}_i$ denote the 3D positions and orientation vectors of corresponding fingertip keypoints for the glove and dex-hand, respectively, balanced by weighting coefficients $\lambda_{\text{pos}}$ and $\lambda_{\text{dir}}$. The term $\mathbf{g}_j$ represents the $j$-th tactile point on the glove, which is mapped to the dex-hand surface via the pre-defined anatomical correspondence function $\mathrm{NN}_{\text{Dex}}(\cdot)$. To prioritize active contact regions, the weight $w_{\mathbf{g}_j}$ is computed from the normalized net force $F_j \in [0,1]$ via the sigmoid activation:
\begin{equation}
    w_{\mathbf{g}_j} = [1 + \exp(-20(F_j - 0.5))]^{-1}
\end{equation}

In addition to retargeting joint angles and end-effector poses, we also synthesize tactile signals for the dexterous hand based on its retargeted configuration. Specifically, given the retargeted joint angles $J_{\text{Dex}}$, the 3D positions of tactile points on dex-hand can be computed through forward kinematic. For each glove tactile point $\mathbf{g}_i$, we identify its anatomically corresponding dex-hand tactile point $\mathbf{q}_i$ via the structural mapping $\mathrm{NN}_{\text{Dex}}(\cdot)$. The spatial discrepancy between the human and robot tactile sites is then quantified by their Euclidean distance:
\begin{equation}
    \delta_i = \left\| \mathbf{g}_i - \mathbf{q}_i \right\|_2
\end{equation}

Intuitively, smaller $\delta_i$ implies better geometric alignment and thus higher fidelity in tactile transfer. We leverage this insight to modulate the original glove tactile signals using a distance-aware attenuation function. Let $\gamma_i$ denote the $i-th$ tactile signal of $\gamma_{Glove}^{t}$. The retargeted tactile signal for the dex-hand is then defined as:
\begin{equation}
    \hat{\gamma}_i = \gamma_i \cdot \frac{1}{1 + \exp\big(-\alpha (\delta_i - \beta)\big)}
\end{equation}
where $\alpha > 0$, $\beta \geq 0$ are tunable hyperparameters controlling the sensitivity and threshold of tactile attenuation. In practice, we set $\beta$ to the expected alignment error under ideal retargeting (e.g., 5–10 mm), and $\alpha$ to sharpen the transition (e.g., $\alpha = 20$).

The resulting retargeted tactile signal $\Gamma_{\text{Dex}} = \{\gamma_{\text{Dex}}^{t}\}_{t=1}^{T}$, where $\gamma_{\text{Dex}}^{t} = \{\hat{f}_i\}_{i=1}^{M}$, prioritizes tactile fidelity over mere geometric alignment. By attenuating tactile responses based on the spatial discrepancy between human and robot fingertips, our method ensures that the contact forces applied to the object remain consistent with human intent. This tactile-aware retargeting is crucial for preserving the physical interaction semantics—such as grip strength and contact timing—thereby enabling reliable execution of force-sensitive manipulation tasks.

By minimizing  $ \mathcal{L} $ at each time stamp, we obtain retargeted joint trajectories and end-effector poses for the dexterous hand. Consequently, the original glove states  $ (J_{\text{Glove}}, P_{\text{Glove}}, \Gamma_{\text{Glove}}) $  are transformed into dex-hand states  $ (J_{\text{Dex}}, P_{\text{Dex}}, \Gamma_{\text{Glove}} )$  via the retargeting operator:
\begin{align}
    P_{\text{Dex}}, J_{\text{Dex}}, \Gamma_{\text{Glove}} &= \mathrm{Retarget}(P_{\text{Glove}}, J_{\text{Glove}}, \Gamma_{\text{Dex}})
\end{align}
where,  $ J_{\text{Dex}} = \{j_{\text{Dex}}^{t}\}_{t=1}^T $ and $ P_{\text{Dex}} = \{p_{\text{Dex}}^{t}\}_{t=1}^T $  denote the joint angles and homogeneous pose matrices (expressed in the RGB-D camera frame) of the dex-hand.

For now, these poses remain anchored in the camera coordinate system and must be transformed into the robot’s operational space. Let  $ M $  denote the calibrated extrinsic transformation from the RGB-D camera frame to the robot base frame. Applying  $ M $  yields robot-centric poses:
\begin{align}
    P_{\text{RobotTCP}} &= M P_{\text{Dex}} \\
    P_{\text{ObjectInRobot}} &= M P_{\text{Object}}
\end{align}
where  $ P_{\text{RobotTCP}}=\{p_{RobotTCP}^t\} $ represents the poses of the robot Tool Center Point (TCP), and  $ P_{\text{ObjectInRobot}}=\{p_{ObjectInRobot}^t\} $  represents the poses of the object relative to the robot base.

\subsection{Rendering Environment-Specific Robot Data}
\subsubsection{Scene Reconstruction}
Faithful background rendering is critical for sim-to-real transfer. To this end, we reconstruct the workspace as a 3D Gaussian Splatting (3DGS) model (\cite{4kerbl3Dgaussians}), which enables real-time, photorealistic novel-view synthesis. 

The reconstructed 3DGS must be aligned with the simulation environment in scale, rotation, and translation. Following the protocol of Re3SIM (\cite{han2025re3simgeneratinghighfidelitysimulation}), we place a planar ArUco marker of known physical size (e.g., 10\,cm) on the worktable during scanning. Let  $ \mathcal{P}^{\text{real}} = \{\mathbf{x}_j^{\text{real}}\} $  denote 3D points of the marker corners measured from real world, and  $ \mathcal{P}^{\text{3DGS}} = \{\mathbf{x}_j^{\text{3DGS}}\} $  their counterparts in 3DGS. We compute a similarity transformation  $ S = (s, R, \mathbf{t}) $  that minimizes:
\begin{equation}
    \min_{s, R, \mathbf{t}} \sum_{j} \left\| s R \mathbf{x}_j^{\text{3DGS}} + \mathbf{t} - \mathbf{x}_j^{\text{real}} \right\|_2^2,
\end{equation}

This closed-form solution ensures metric alignment between real and virtual environments. Once aligned, the 3DGS model is exported as a Universal Scene Description (USD) asset and imported into Isaac Sim, serving as the static background for dynamic rendering. This step guarantees geometric and photometric consistency, thereby reducing the domain gap in visual observations.

\subsubsection{Robot Data Rendering}

As demonstrated in recent literature, visual input plays a pivotal role in the success of embodied intelligence models. High-fidelity visual rendering significantly enhances policy learning and task execution accuracy by providing richer geometric and semantic cues. To this end, we employ a hybrid rendering scheme in Isaac Sim 5.1, where the static background environment is rendered using 3D Gaussian Splatting (3DGS) for photorealism, while dynamic objects and robotic manipulators are rendered using high-quality mesh models.

The entire rendering environment's coordinate system is aligned with the robot base frame. Specifically, at each time step $t$, the joint angles of the robot arms are computed using an inverse kinematics (IK) solver based on the target pose matrices $p_{\text{RobotTCP}}^t$ of the dex-hand. The resulting joint angles drive the robot arms and their end-effectors, while object poses $p_{\text{ObjectInRobot}}^t$ drive the motion of manipulated objects at the same time.

To achieve high-quality rendering of dynamic components, we utilize the path-tracing rendering engine provided by Isaac Sim. This ensures realistic lighting and shadow effects, crucial for simulating complex interactions between the robot and its environment. At each time step, the visual scene is composed by overlaying the dynamically rendered robot and objects onto the static 3DGS background. This approach not only guarantees photorealism but also supports flexible image editing from arbitrary viewpoints and robot embodiments.

Finally, the rendered visual images and other modalities are packaged into VLA training datasets. Specifically, at time stamp $t$, the robot action data $a^t$ is defined as:
\begin{equation}
    a^t = [pos^t, rot^t, j_{Dex}^t] 
\end{equation}
Where $pos^t, rot^t \in \mathbb{R}^3$ represent the translation and rotation direction vector derived from the robot TCP pose $P_{\text{RobotTCP}}$, respectively, and $j_{Dex}^t \in \mathbb{R}^N$ denotes the dex-hand joint angles. In addition to standard action and visual data, we optionally incorporate tactile feedback processed as tactile images following the ObjTac format (\cite{cheng2025omnivtlavisiontactilelanguageactionmodelsemanticaligned}). The integrated training data $d^t$ is then represented as:
\begin{equation}
    d^t = [a^t, img_{\text{visual}}^t, (img_{\text{tactile}}^t)] 
\end{equation}
Here, $img_{\text{visual}}^t$ denotes the rendered visual observation described previously, while $img_{\text{tactile}}^t$ is the processed tactile heatmap capturing the contact forces and pressures experienced by the robot's fingertips. This comprehensive dataset provides a robust foundation for training VLA models.

\section{Experiments}

We evaluate our method from two aspects: simulation evaluation and real-world evaluation on Vision-Language-Action (VLA) models. In simulation evaluation part, we visually evaluate the precision of pose estimation and tactile signal obtained from our data acquisition pipeline. Besides, we visually evaluate the fidelity of the edited data compared with the real data. As for the downstream performance of Vision-Language-Action (VLA) models trained on our data, we conduct three different tasks for evaluation.

\begin{figure*}[t]
    \centering
    \includegraphics[width=1\textwidth]{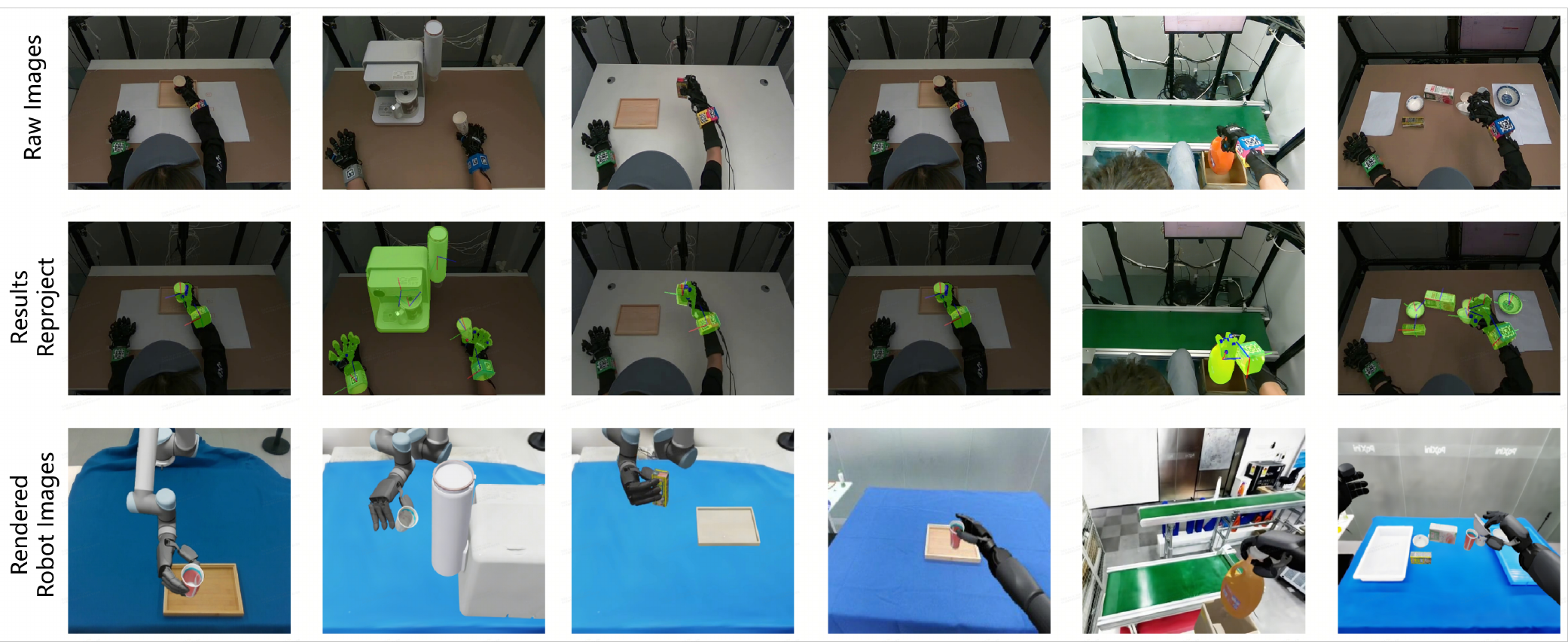}
    \caption{Simulation validation of our data collection pipeline and Real-Sim-Real data processing pipeline. The first row is the raw images of collected data. The second row is the re-projection results of glove, object and tactile point. The last row is the rendered images using our Real-Sim-Real data pipeline.}
    \label{fig:sim_exp}
\end{figure*}

\begin{figure*}[t]
    \centering
    \includegraphics[width=1\linewidth]{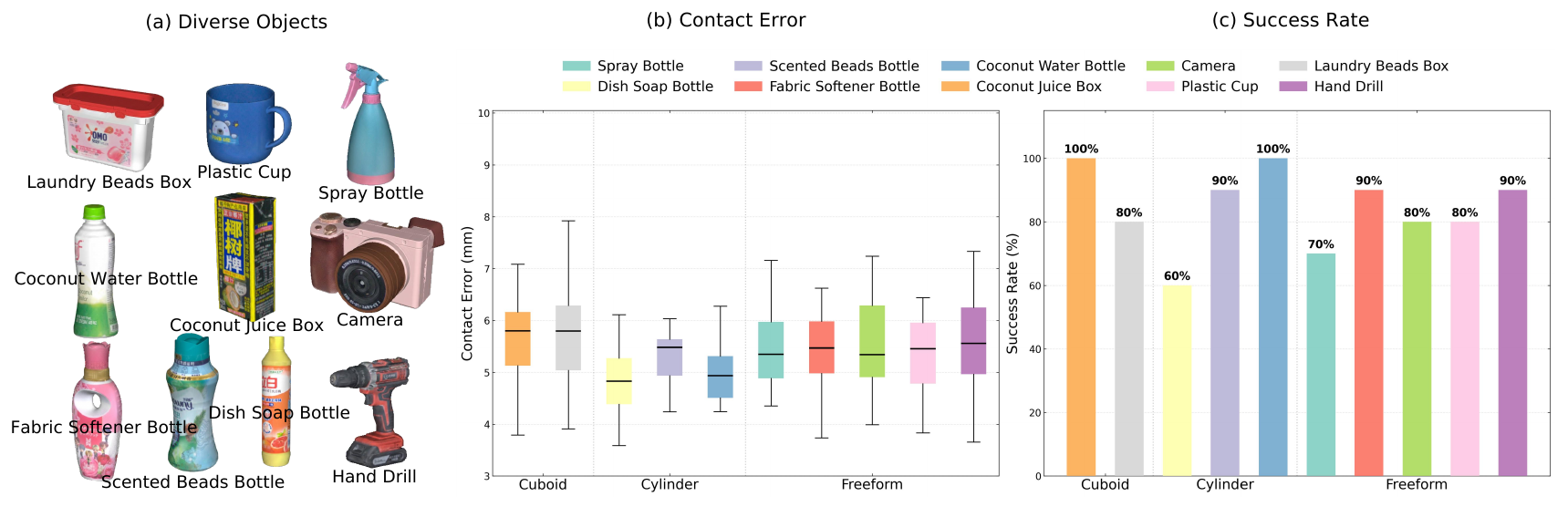}
    \caption{Contact error in the simulation enviroment and success rates of real-world replay for different objects. Each object was tested on 10 demonstrations using retargeted DexH13 joint angles and retargted dex-hand end trajectories.}
    \label{fig:contact-error}
\end{figure*}

\subsection{Simulation Evaluation}\label{subsec4}


To verify the precision of our originally collected human demonstration data, we perform a re-projection analysis in the image space. Specifically, based on the collected multi-modal data, we can compute the wrist pose and object pose under world coordinate system. As a result, 3D glove can be reconstructed using joint angle and glove URDF. Furthermore, the tactile contact points can be computed by mapping the sensor coordinate into world coordinate via forward kinematics of the instrumented glove. With all of above information, we re-project the estimated 3D glove, 3D objects, and tactile contact points (annotated with red-to-blue point according to its contact force) back onto the corresponding RGB frames using the intrinsic and extrinsic camera parameters.

The raw image and the re-projected results are shown in the first and second rows of Fig.\ref{fig:sim_exp}, respectively. Subjective inspection across hundreds of frames shows tight alignment between re-projected 3D gloves and their physical counterparts, as well as precise overlay of object meshes onto real-world objects. Crucially, the reprojected tactile contact points consistently coincide with visible regions of finger-object interaction (e.g., fingertips pressing against a mug handle or tool surface), and the magnitude of the rendered force correlates with observed deformation or grip intensity. This cross-modal consistency confirms the temporal synchronization and geometric accuracy of our multimodal data pipeline, establishing a reliable foundation for subsequent retargeting and simulation.

Moreover, these raw data are further processed by our Real-Sim-Real pipeline using two robot embodiment: UR5 and Paxini ToRA One. The final generated visual images are presented in the last row of Fig.\ref{fig:sim_exp}. It can be seen that our method generates photorealistic robot demonstration data.

To further validate the accuracy of our proposed Dex-Tactile retargeting method, we evaluate retargeting accuracy across 10 diverse objects with varying shapes. The 10 objects are shown in Fig.\ref{fig:contact-error} (a). For each object, by importing both the high-fidelity object and the corresponding retargeted dex hand, we replay the retargeted manipulation process in Isaac Sim. During replay, the original glove tactile contact points are mapped into the tactile contact points of the dex hand using our anatomical correspondence function $\mathrm{NN}_{\text{Dex}}(\cdot)$. We then calculate the Euclidean distance between these mapped points and the actual simulated contact locations on the dex-hand surface. The results over 10 diverse objects are reported in Fig.\ref{fig:contact-error} (b). The average error of tactile contact points is 3.86 mm. The results confirm that our dex-tactile retargeting method produces dex hand actions that are both geometrically valid and interaction-consistent.

\subsection{Real-World Evaluation}\label{subsec5}

\begin{table*}[t]
\centering
\caption{Real-world experimental results on different models using different data.}
\label{tab:hyper_vtla_vta}
\tabcolsep=5mm
\begin{tabular}{c|cccccc}
\toprule
\textbf{Task} & \multicolumn{2}{c}{\textbf{DP}} & \multicolumn{2}{c}{\textbf{Pi05 (w/ wrist camera)}} & \multicolumn{2}{c}{\textbf{Pi05 (w/o wrist camera)}} \\
& Tele & Paint & Tele & Paint & Tele & Paint  \\
\hline
Pick and place       & 90.0\%  & 40.0\% &  100.0\% & 70.0\% & 90.0\% & 40.0\%   \\
Push cuboid           & 100.0\% & 100.0\% &  100.0\% & 100.0\% & 90.0\% & 70.0\%   \\
Pour bottle           & 40.0\%  & 10.0\% &  100.0\% & 70.0\% & 70.0\% &  30.0\%  \\
\rowcolor[HTML]{EFEFEF} 
\textbf{Avg.}    & \textbf{76.6\%}  & \textbf{50.0\%} &  \textbf{100.0\%} & \textbf{80.0\%} & \textbf{83.3\%} &  \textbf{46.6\%} \\   
\bottomrule
\end{tabular}
\end{table*}

The performance of our Dex-Tactile retargeting method is further validated in real world. We conduct replay experiments on UR5 equipped with the Paxini DexH13 dex-hand. For each of the retargeted demonstration, UR5 end-effector is driven according to the retargeted trajectory and dex-hand is driven according to the retargeted joint angles. The object is placed at the same position when human demonstration data collected, ensuring consistent starting conditions.

We perform 10 independent trials per object. A trial is considered successful if the robot completes the intended manipulation task, and achieves the desired physical outcome (e.g., lifting, pouring, or pushing). The success rates across all objects are summarized in Fig.\ref{fig:contact-error} (c).
As shown, the robot successfully completes most tasks, achieving an average success rate of 84$\%$. Notably, high success rates (90–100) are observed for objects with simple geometry and stable contact modes (e.g., coconut water bottle, fabric softener bottle), indicating that our retargeting method effectively preserves grasp stability and motion feasibility. For more complex objects such as the plastic cup and camera, the success rate remains above 80, demonstrating robustness under geometric variability. These results confirm that our retargeted trajectories are not only geometrically plausible in simulation but also physically executable on real hardware. More importantly, these experiments validate that incorporating tactile feedback into the retargeting process enhances interaction fidelity, enabling reliable execution of dexterous manipulation tasks in the real world.

To comprehensively evaluate the efficacy of our Real-Sim-Real data pipeline for robot skill learning, we conduct a comparative study by training Vision-Language-Action (VLA) policies using two distinct sources of demonstration data: (1) synthetic data generated through our proposed pipeline (referred to as Real-Sim-Real data), and (2) real-world teleoperation demonstrations collected via direct human control of the robot (referred to as Tele. data). To ensure a rigorously fair comparison and isolate the impact of the data source, both datasets are strictly balanced to contain an equal number of demonstrations across three representative manipulation tasks: pick place, pushing a cube, and \textit{pouring water from a bottle}. These tasks were specifically selected to evaluate the policy's robustness across varying degrees of contact dynamics and precision requirements.

We choose Pi0.5 (\cite{intelligence2025pi05visionlanguageactionmodelopenworld}) and DP (\cite{chi2024diffusionpolicyvisuomotorpolicy}) as the VLA architecture for evaluation. All policies are trained using the same hyperparameters across different types of data mentioned above. During deployment, each policy is evaluated on 10 randomly sampled initial object positions within the robot’s workspace. A trial is considered successful if the robot completes the task without human intervention and achieves the desired physical outcome, i.e., the target object is placed in the goal region, the cube is pushed to the designated location, and the bottle is tilted by nearly 90 degrees.

The success rates are summarized in Tab.\ref{tab:hyper_vtla_vta}. While policies trained on teleoperation data achieve 100\% success rate over these tasks, those trained on our Real-Sim-Real data achieve 80\% success rate. There only exhibits 20\% drop in success rate. These findings suggest that our pipeline can serve as a scalable and cost-effective alternative to teleoperation for generating high-quality training data, particularly when teleoperation is impractical or costly.

\begin{table*}[t]
\caption{Efficiency of different data collection methods.}
\begin{tabular}{c|cccccc}
\hline
               & pick and place & open box    & push cubiod & bagging fruits & table bussing & fold clothes \\ \hline
Teleoperation & $\sim$1h30min  & $\sim$2h    & $\sim$2h    & $\sim$10h              & $\sim$12h     & $\sim$16h    \\
Human          & $\sim$35min    & $\sim$30min & $\sim$30min & $\sim$2h20min          & $\sim$2h30min       & $\sim$3h     \\
\rowcolor[HTML]{EFEFEF} 
\textbf{Efficiency rate} & \textbf{2.57}           & \textbf{4.00}           & \textbf{4.00}           & \textbf{4.28}                   & \textbf{4.80}           & \textbf{5.33}         \\ \hline
\end{tabular}
\label{tab:data_efficiency} 
\end{table*}

\subsection{Data Collection Efficiency}
\label{subsec:efficiency}
A critical advantage of our human demonstration pipeline is its superior data collection efficiency compared to direct robot teleoperation, particularly for complex dexterous tasks. To quantify this, we collected 100 successful demonstration sequences across 6 tasks of varying difficulty using both methods. All trials were conducted by trained operators, where a demonstration is defined as successful if the task is completed without object drops or resets.

As summarized in Table \ref{tab:data_efficiency}, human demonstration consistently outperforms teleoperation across all tasks. For the basic pick-and-place task, teleoperation requires $2.57\times$ the time relative to human collection. This gap widens significantly as task complexity increases. Notably, in the multi-stage table bussing and folding clothes tasks, our method collects the same amount of valid data in approximately one-fifth the time required by teleoperation (up to $5.33\times$ speedup), as these tasks demand precise bimanual coordination and delicate force control.

\section{Conclusion and Future Work}

We present a scalable human demonstration data collection framework that captures synchronized visual, kinematic, and tactile signals using self-developed gloves under calibrated multi-view monitoring environment. Based on the data collection framework, we develop a Real-Sim-Real data processing pipeline that retargets human demonstrations to robot embodiments with tactile-aware optimization and renders photorealistic training data via metrically aligned 3D Gaussian Splatting..

We believe our pipeline enables the generation of semantically rich, cross-embodiment robot datasets. Experiments show that policies trained on our Real-Sim-Real data achieve high success rate in real-world dexterous tasks. We will release our data processing toolkit\footnote{\url{https://github.com/px-DataCollection/px_omnisharing_dataprocess_kit}} to support community research.

\bibliographystyle{assets/plainnat}
\bibliography{paper}

\clearpage
\newpage
\section{Supplementary Materials}

\subsection{Data Augmentation}
To further enhance the diversity and robustness of the training dataset, we also incorporate data augmentation functionalities. This technique not only simulates a wider range of operational scenarios but also effectively reduces the risk of model overfitting.

(1) Background Swap. By capturing and reconstructing diverse desktop manipulation scenes, we have compiled a collection of background assets for utilization. Therefore, the generated data can be duplicated under various background settings, making the learned robotic behaviors more versatile and independent of specific background conditions.

(2) Object Material Swap. In addition to the original texture of the objects, we have also generated many other textures using open-source 3D generation toolkit. Similarly, the manipulated objects can be rendered under different textures. Such variations are also crucial for training robots to recognize and adapt to objects of different materials.

\begin{figure}[h]
    \centering
    \includegraphics[width=0.9\textwidth]{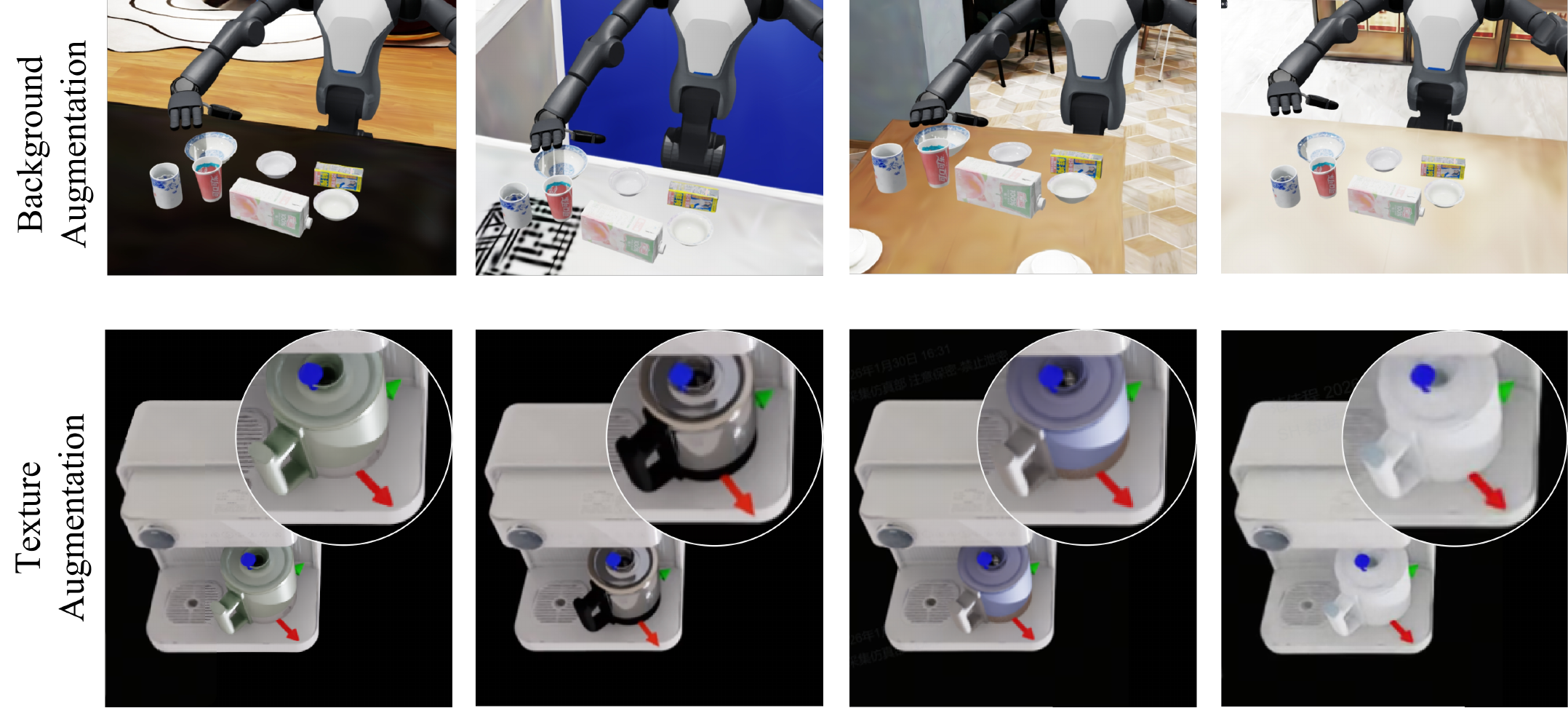}
    \caption{Data augmentation examples showing background replacement and object material changes. Top row shows the original scene, middle row demonstrates background change, and bottom row illustrates object material variation.}
    \label{fig:aug}
\end{figure}

To better illustrate this process, we provide an example figure (Fig. \ref{fig:aug}), demonstrating the changes of  background and object materials. In this figure, the images on the top row displays the swap of static background, the images on the swap of object material. Through this approach, we can significantly enrich the content of the training dataset, enhancing the adaptability and practicality of the model.

\subsection{Model Deployment Results}

\begin{figure*}[t]
    \centering
    \includegraphics[width=1\linewidth]{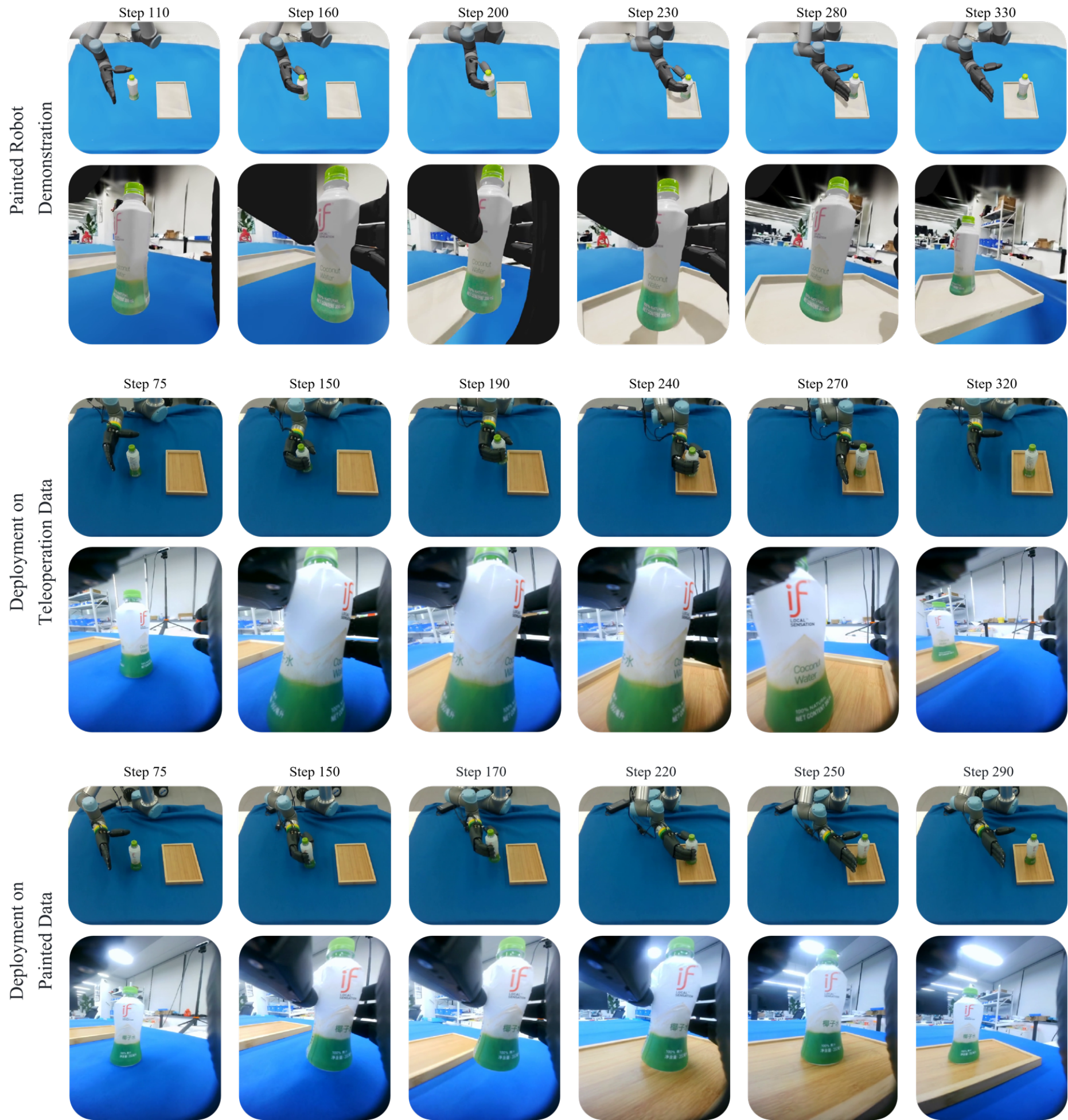}
    \caption{Deployment results for Task 1 (Pick and Place). The top two rows show the "painted" images from our Real-Sim-Real pipeline. Rows 3 and 4 show the deployment results using teleoperated data, while rows 5 and 6 show the deployment results using Real-Sim-Real data.}
    \label{fig:deploy_results_task1}
\end{figure*}

\begin{figure*}[t]
    \centering
    \includegraphics[width=1\linewidth]{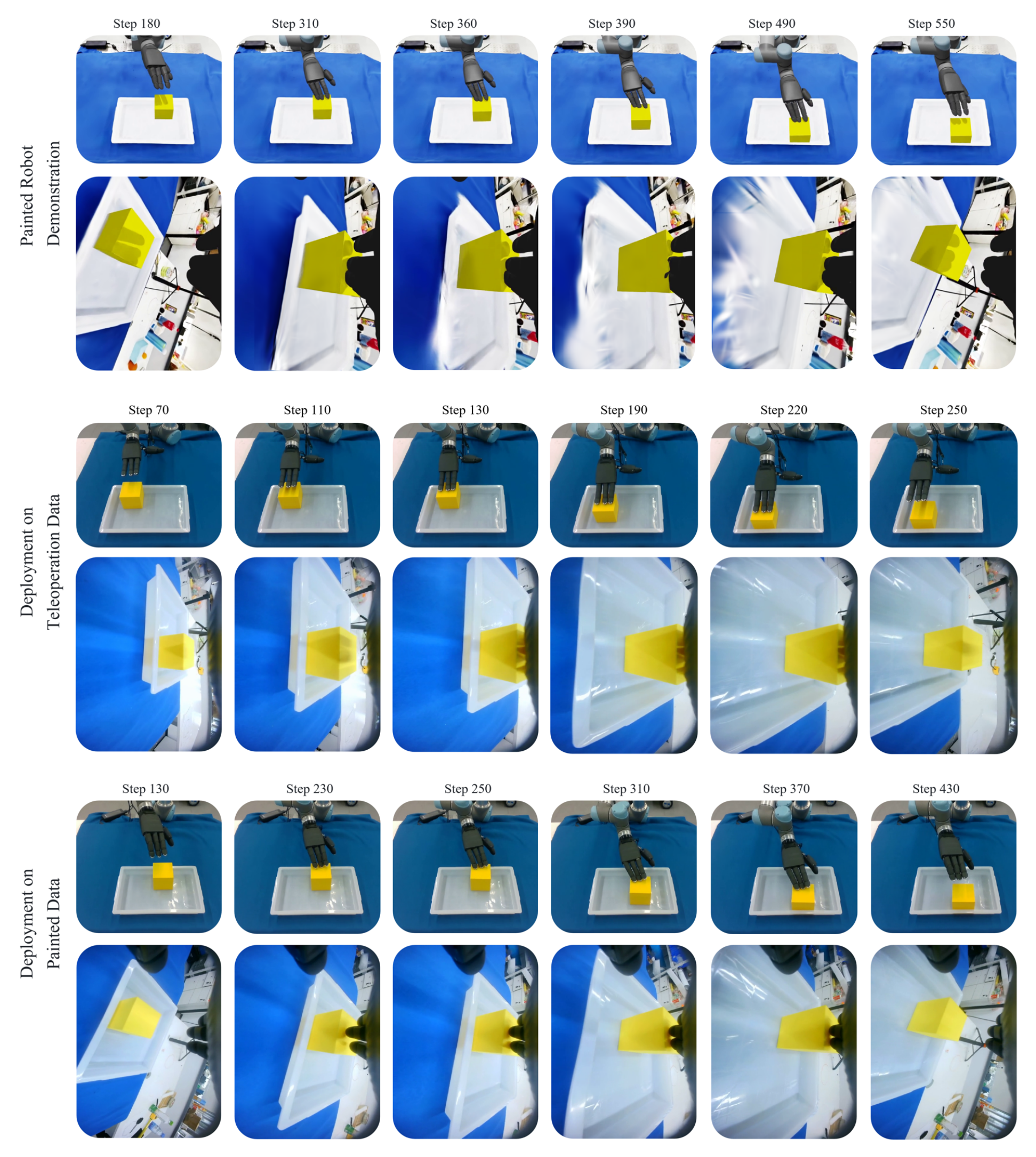}
    \caption{Deployment results for Task 2 (Push Cuboid). The top two rows show the "painted" images from our Real-Sim-Real pipeline. Rows 3 and 4 show the deployment results using teleoperated data, while rows 5 and 6 show the deployment results using Real-Sim-Real data. Note that the "painted" wrist images exhibit some artifacts. This is because the wrist camera is too close to the 3DGS point cloud, causing the relevant 3D points to be culled by the renderer. We will optimize the 3DGS reconstruction algorithm to improve the quality of the 3DGS reconstruction. }
    \label{fig:deploy_results_task2}
\end{figure*}

\begin{figure*}[t]
    \centering
    \includegraphics[width=1\linewidth]{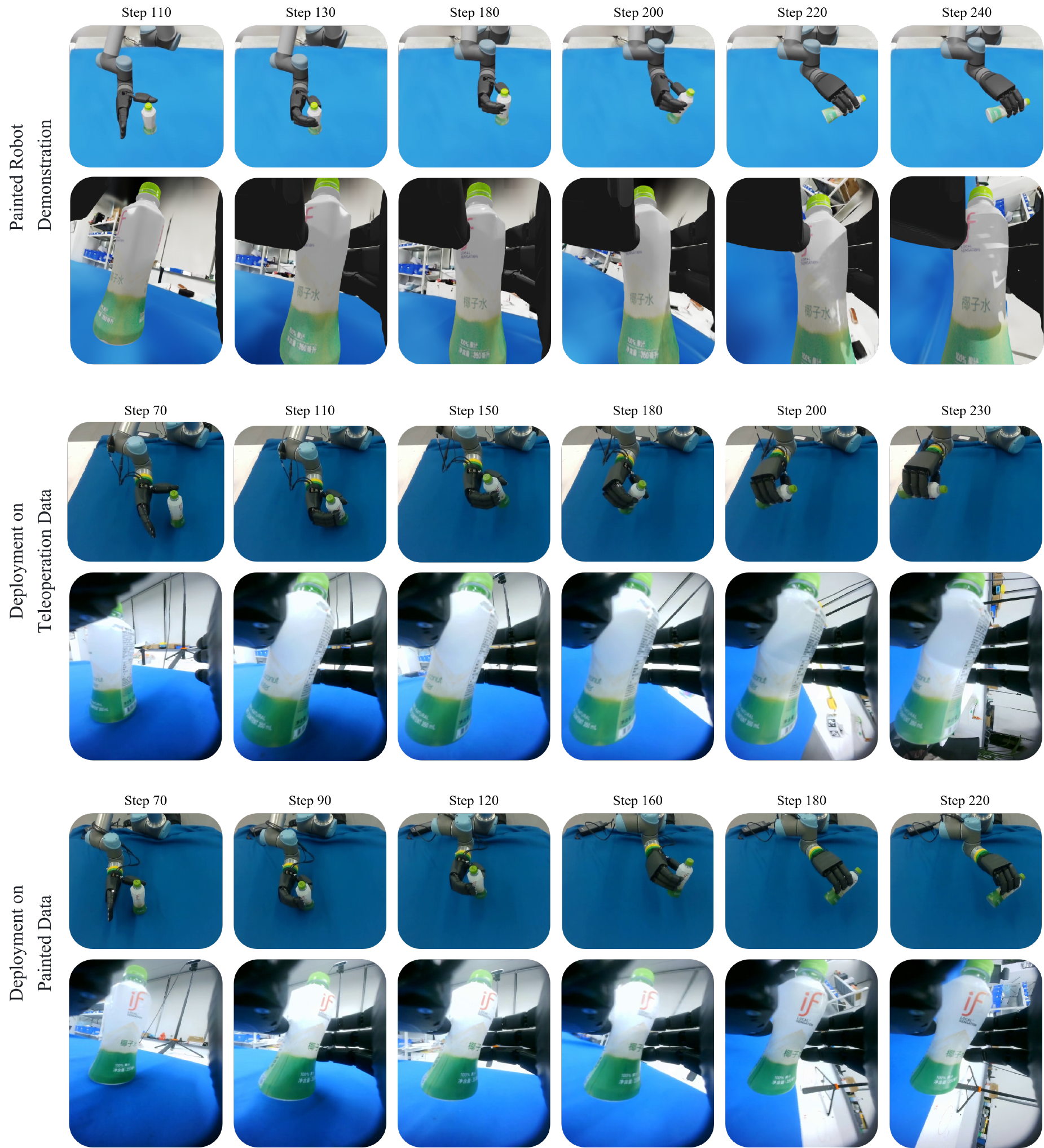}
    \caption{Deployment results for Task 3 (Pour). The top two rows show the "painted" images from our Real-Sim-Real pipeline. Rows 3 and 4 show the deployment results using teleoperation data, while rows 5 and 6 show the deployment results using Real-Sim-Real data.}
    \label{fig:deploy_results_task3}
\end{figure*}

We present the deployment results of Pi0.5 over the three different manipulation tasks mentioned in Experiments Section. As shown in Fig.\ref{fig:deploy_results_task1}, Fig.\ref{fig:deploy_results_task2} and Fig. \ref{fig:deploy_results_task3}, each figure is organized into six rows. The first two rows show the "painted" main images and wrist images obtained from our Real-Sim-Real pipeline. The third and fourth rows display the main images and wrist images obtained during the deployment of Pi0.5, which trained on teleoperation data. The fifth and sixth rows display the main images and wrist images obtained during the deployment of Pi0.5, which trained on Real-Sim-Real data.

\begin{figure*}[t]
    \centering
    \includegraphics[width=1\linewidth]{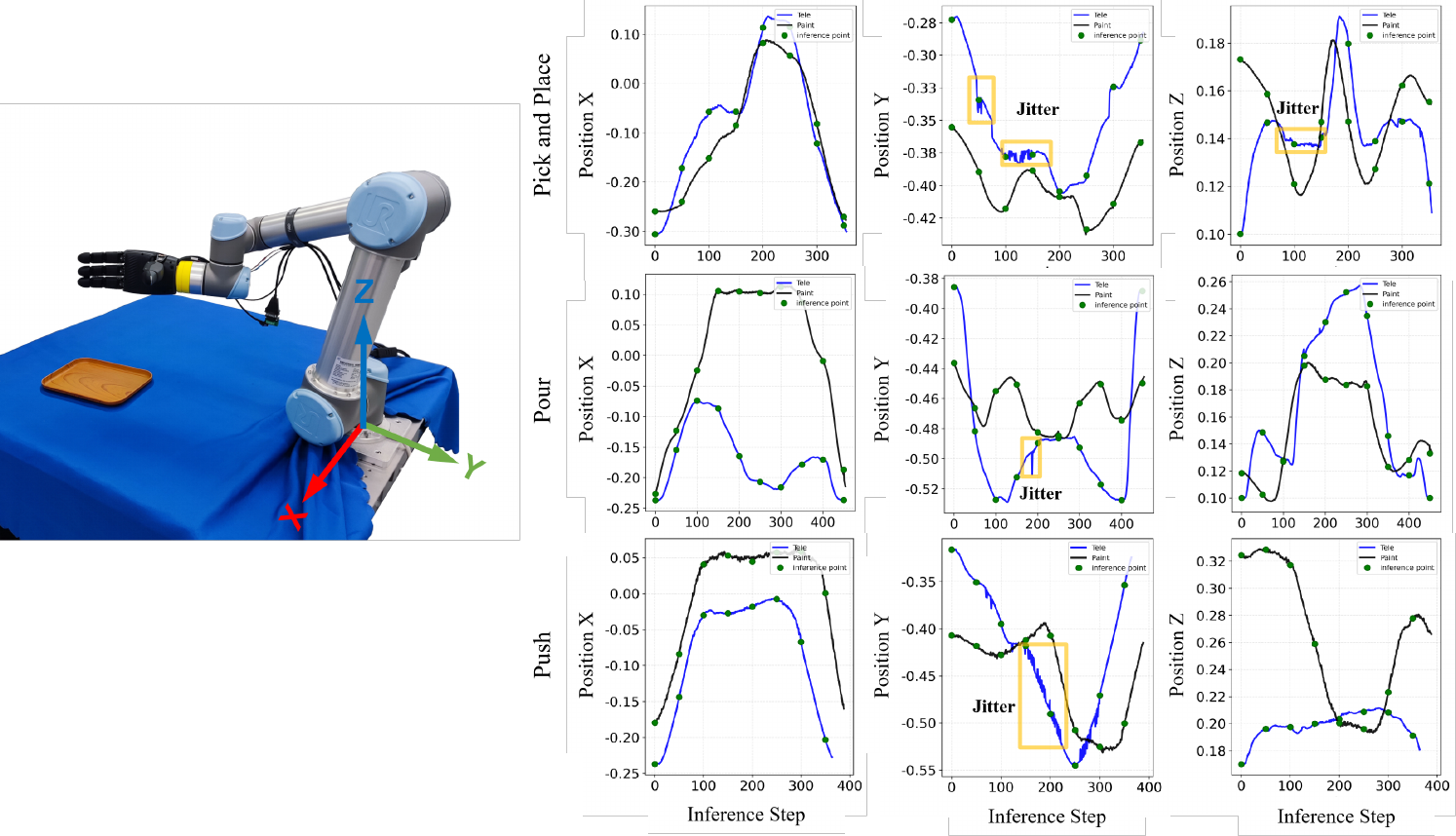}
    \caption{Deployment action curves of Pi0.5 training on different data. The black curve shows the deployment action curve using our Real-Sim-Real data as training data. The blue curve shows the deployment action curve using teleoperation data as training data. Note that since the sources of teleoperation and human-collected data differ in the physical enviroment, so the initial hand positions were not strictly constrained as the same position. However, the trajectories demonstrate that compared to teleoperation, the \textbf{RoboPaint} trajectory aligns more closely with human intuitive operation and reduces stiff movements.}
    \label{fig:deploy_traj}
\end{figure*}

Furthermore, we visualize and analyze the deployment action trajectories of models trained on human demonstrations versus those trained on teleoperation data. As shown in Fig.\ref{fig:deploy_traj}, models trained on our Real-Sim-Real data exhibit smoother and more stable execution during deployment, whereas those trained on teleoperation data display noticeable jitter. We attribute this discrepancy to fundamental limitations in the teleoperation data acquisition pipeline: the system must simultaneously track fine-grained robot states and relay human intent through intermediate controllers (e.g., joysticks or motion capture suits), which inherently introduces a trade-off between pose estimation accuracy and trajectory smoothness. In practice, small tracking errors, sensor noise, or control latency accumulate over time, resulting in non-smooth action sequences that the policy learns as part of the demonstration distribution. In contrast, our human demonstration approach captures natural, continuous manipulation motions without intermediary actuation artifacts, yielding cleaner training signals that promote temporally coherent policy outputs.




\end{document}